\documentclass[10pt,twocolumn,letterpaper]{article}
\usepackage[pagenumbers]{cvpr} 
\usepackage{gensymb}
\usepackage{array,multirow}
\usepackage{mwe}
\usepackage[table]{xcolor}
\usepackage{algpseudocode}
\usepackage{algorithm}
\usepackage{booktabs}
\usepackage[table]{xcolor}
\usepackage{colortbl}
\usepackage{tikz}
\usepackage{graphicx}
\usepackage{stmaryrd}
\usepackage[normalem]{ulem}
\usepackage{tcolorbox}
\usepackage{mathtools}
\usepackage{dsfont}
\usepackage{etoolbox}

\definecolor{cvprblue}{rgb}{0.21,0.49,0.74}
\usepackage[pagebackref,breaklinks,colorlinks,citecolor=cvprblue]{hyperref}

\def\paperID{10518} 
\def\confName{CVPR}
\def\confYear{2024}
\def\onedot{\ifx\@let@token.\else.\null\fi\xspace}

\def\ie{\emph{i.e}\onedot} 
 
 \def\vs{\emph{vs}\onedot}
 
\def\etal{\emph{et al}\onedot}
 
\makeatother

\def\Vec#1{{\boldsymbol{#1}}}

\newcommand{\insertfig}{\includegraphics[width=\textwidth]{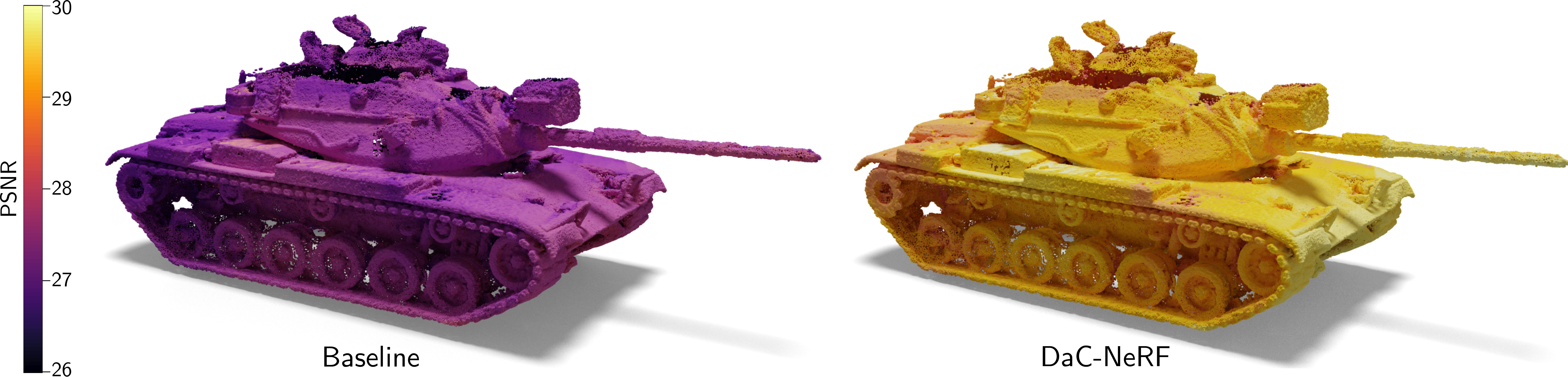}
\captionof{figure}{Rendering quality comparison. We conduct a rendering quality evaluation by measuring the Peak Signal to Noise Ratio (PSNR) between rendered views of the specialized models trained using our proposed Divide and Conquer (DaC) training pipeline and one trained through the standard training pipeline which assumes equal importance for all training set images. The PSNR values are projected onto the 3D point cloud of the rendered scene. This comparison highlights that through spatial specialisation, the DaC training paradigm allows for a superior learning of the 3D scene.}
\label{fig:top_fig}
\vspace{3em}}

\usepackage{pifont}
\newcommand{\redColorMap}[1]{
    \ifdim #1 pt>36pt
        \cellcolor{red!90}
    \else
        \ifdim #1 pt>35pt
            \cellcolor{red!70}
        \else
            \ifdim #1 pt>34pt
                \cellcolor{red!50}
            \else
                \ifdim #1 pt>33pt
                    \cellcolor{red!30}
                \else
                    \cellcolor{red!10}
                \fi
            \fi
        \fi
    \fi
}

\title{Divide and Conquer: Rethinking the Training Paradigm of Neural Radiance Fields}

\author{Rongkai Ma$^1$\thanks{Corresponding author}~~\thanks{Work done while at CSIRO}\quad Leo Lebrat$^2$\quad Rodrigo Santa Cruz$^2$\quad Gil Avraham$^3$ \\ Yan Zuo$^3$\quad Clinton Fookes$^4$\quad Olivier Salvado$^2$\\\\
$^{1}$Nvidia,\quad $^{2}$CSIRO, Data61,\quad $^{3}$Amazon,\quad $^{4}$Queensland University of Technology\\
}

\makeatletter
\apptocmd{\@maketitle}{\centering\insertfig}{}{}
\makeatother

\begin{document}


\maketitle
\begin{abstract}
Neural radiance fields (NeRFs) have exhibited potential in synthesizing high-fidelity views of 3D scenes but the standard training paradigm of NeRF presupposes an equal importance for each image in the training set. This assumption poses a significant challenge for rendering specific views presenting intricate geometries, thereby resulting in suboptimal performance. In this paper, we take a closer look at the implications of the current training paradigm and redesign this for more superior rendering quality by NeRFs. Dividing input views into multiple groups based on their visual similarities and training individual models on each of these groups enables each model to specialize on specific regions without sacrificing speed or efficiency. Subsequently, the knowledge of these specialized models is aggregated into a single entity via a teacher-student distillation paradigm, enabling spatial efficiency for online rendering. Empirically, we evaluate our novel training framework on two publicly available datasets, namely NeRF synthetic and Tanks\&Temples. Our evaluation demonstrates that our DaC training pipeline enhances the rendering quality of a state-of-the-art baseline model while exhibiting convergence to a superior minimum.

\end{abstract}    
\section{Introduction}
\label{sec:intro}
Rendering high-quality images of real-world scenes remains a significant challenge, particularly those with complex geometries. Neural Radiance Fields (NeRFs~\cite{mildenhall2020nerf}) represent a major leap in producing detailed views using volumetric rendering. 
However, NeRF's conventional training approach assigns the same importance to all scene perspectives available for training. It uniformly compresses the geometric and photometric information of the scene into the neural network weights. This approach tends to disregard the natural asymmetry of details present in diverse perspectives of complex scenes, leading to declined rendering quality for synthesizing novel views~\cite{pan2022activenerf}. In a scenario of a flat surface having intricate details on just one side, we expect that perspectives depicting the detailed faces would hold greater significance than those from the uniformly flat side.

Various works have been devised to improve the limitations of NeRFs, including improved space sampling~\cite{barron2021mip, barron2022mip, barron2023zipnerf}, explicit spatial feature learning~\cite{chen2022tensorf,cao2023hexplane,fridovich2023k}, and multi-resolution hashable grid techniques~\cite{muller2022instant} to enhance scene representation. While these methods show improved rendering, they still rely on standard training techniques. We differentiate our work by a novel divide and conquer approach, which is a generic training pipeline and can be applied more universally to any type of NeRF models. We draw inspiration from ensemble learning~\cite{breiman1996bagging,smith1985design} and mixture of experts (MoE) concepts~\cite{yuksel2012twenty,riquelme2021scaling,zhenxing2022switch}, where multiple models are trained on different scene partitions (known as bootstrapping) and are aggregated during inference. Whilst this technique has seen great success in large-scale NeRF scene reconstruction~\cite{zhenxing2022switch,tancik2022block}, they fall short when managing computational resources during the inference stage.

\begin{figure}[t]
  \begin{subfigure}[]{0.49\columnwidth}
    \includegraphics[width=\linewidth]{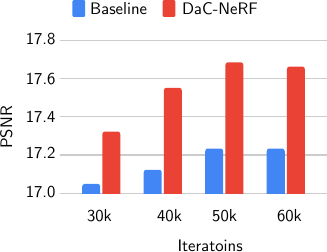}
    \caption{Train}
    \label{subfig:1}
  \end{subfigure}
  \hfill 
  \begin{subfigure}[]{0.49\columnwidth}
    \includegraphics[width=\linewidth]{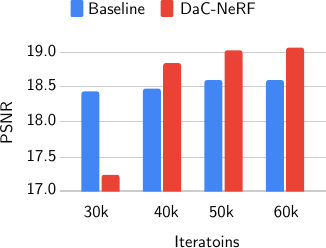}
    \caption{M60}
    \label{subfig:2}
  \end{subfigure}
  \caption{Convergence comparison. We conduct the evaluation of the model trained with our proposed DaC paradigm and the standard pipeline, respectively, on the Train (a) and M60 (b) scenes at $30,000$, $40,000$, $50,000$, and $60,000$ iterations. The results demonstrate that the model trained with the DaC's experts reaches a superior convergence point compared to the standard training strategy.} 
  \label{fig:distilled_baseline}
\end{figure}

\textbf{Our work.} We present a novel training pipeline for neural radiance fields in addressing the complexities of intricate scenes beyond the capabilities of a single NeRF model via a \underline{D}ivide \underline{a}nd \underline{C}onquer (DaC) approach. We initialize our method by dividing the input views into multiple groups based on their inherent structural characteristics.
Subsequently, we train an individual expert NeRF model for each group of views, allowing for a more specialized and refined reconstruction of the scene. This can be illustrated in \cref{fig:top_fig}, where the PSNRs of our specialised models (\ie, experts) and the baseline (\ie, NeRF trained on the full dataset) are projected onto the corresponding region of the 3D model. The visualized results distinctly showcase the superior rendering performance of the ensemble of expert models against a single model. Moreover, these expert models are independent and can thus be trained in parallel, significantly reducing the time complexity of such an approach with parallel training. At the core of our method, we leverage teacher-student distillation~\cite{hinton2015distilling} to aggregate the knowledge from these experts into one unified model. This approach preserves the original capacity of the NeRF model under consideration without introducing additional time complexity during inference.

Our numerical experiments demonstrate that DaC outperforms the standard NeRF training strategy, particularly in overcoming their novel view rendering performance stagnation. For example, in Figure~\ref{fig:distilled_baseline}, the performance of the K-Planes NeRF model on the Tanks and Temples dataset tends to plateau after 40k standard training iterations. In contrast, employing the DaC paradigm enables continued improvement in novel view rendering performance beyond this point. Notably, DaC can be applied to most existing NeRF models without introducing runtime or memory overhead during inference. These features make DaC a generic and powerful approach for advancing the next generation of NeRF-based technologies.


\section{Related work}
\label{sec:related}
The NeRF~\cite{mildenhall2020nerf} literature is broad and diverse. However, the works closely related to ours can be categorized into two groups: methods for improving NeRF's training and rendering efficiency, and approaches involving scene partitioning to create specialized NeRF models for individual partitions. In the following paragraphs, we discuss methods falling within these categories. 

\paragraph{Improving speed and quality of NeRFs}
To improve the training efficiency of a NeRF, a line of research~\cite{reiser2021kilonerf,muller2022instant,cao2023hexplane,chen2022tensorf,fridovich2023k,fridovich2022plenoxels,yu2021plenoctrees, liu2020neural,hu2023tri,hedman2021baking} propose to replace the coordinate-based multi-layer perceptron adopted in original the NeRF~\cite{mildenhall2020nerf} with a sparse (sometimes decomposed) voxel grid of features to represent the radiance field. Specifically, TensoRF~\cite{chen2022tensorf} propose to model the scene by a factorization of the sparse voxel grid into multiple low-rank components. K-Planes~\cite{fridovich2023k} similarly achieves efficiency via decomposing the voxel grid into 3 orthogonal feature planes for static scenes. Thanks to the orthogonal decomposition, K-Planes can represent spatial-temporal features by incorporating an additional temporal axis to model dynamic scenes. However, the trade-off between quality and efficiency is inevitable~\cite{hu2023tri}. 
Tancik~\etal~\cite{tancik2021learned} propose to mitigate this trade-off by employing a more generalized initialization via meta-learning~\cite{mildenhall2020nerf,finn2017model}, enabling not only faster convergence but also enhancing rendering quality. While our proposed method can be categorized within the same family, our initialization distinguishes itself by distillation from multiple experts and is not dependent on inner-loop optimization during the inference stage, resulting in faster rendering. Moreover, NeRFLiX~\cite{zhou2023nerflix} enhances rendering quality by learning the reverse degradation process generated by the radiance field, utilizing two neighbouring views and a NeRF-Sytle Degradation Simulator (NDS). 
Another line of research~\cite{barron2021mip,barron2022mip,barron2023zipnerf,hu2023tri} is committed to alleviating the aliasing effect induced by NeRF-style sampling via rendering conical frustum instead of rays. Our method can be utilized seamlessly and incorporated into these methods to boost model performance further.

\paragraph{Partitioning Scene of Neural Radiance Field} 
The practice of partitioning large-scale scenes and training a dedicated neural radiance field for each partition is reminiscent of ensemble learning~\cite{breiman1996bagging}. This approach explored recently~\cite{zhenxing2022switch,tancik2022block,turki2022mega,zhang2023efficient}. Specifically, BlockNeRF~\cite{tancik2022block} excels in faithfully reconstructing the entire San Francisco city by combining predictions from NeRF models trained on individual city blocks. MegaNeRF~\cite{turki2022mega} leverages a clustering method to partition the scene into a top-down 2D grid which facilitates data parallelism for fast training. Switch-NeRF~\cite{zhenxing2022switch} takes a step further by replacing these heuristic hand-crafted decomposition procedures with an end-to-end learnable gated mixture of experts function, which learns to dispatch a 3D-point to the corresponding specialized NeRF model. Although these methods offer a feasible solution for large-scale scene representation, the substantial memory complexity incurred poses a significant constraint when deploying numerous models into limited computational resources. In contrast, our proposed ``divide and conquer" strategy provides a prospective solution for this online memory constraint by consolidating knowledge from multiple specialized models into a unified one using student-teacher distillation~\cite{hinton2015distilling}. 

Although our distillation strategy exhibits some resemblances to kiloNeRF~\cite{reiser2021kilonerf}, the fundamental objective and formulation of our approach diverge markedly. KiloNeRF distills the knowledge of a solitary NeRF trained on the entire scene, directing it towards lower-capacity, space-constrained nerfs to mitigate rendering artifacts in their collaborative rendering process. In contrast, our distillation methodology aims to aggregate knowledge from multiple NeRF experts into a single NeRF model. This consolidated model surpasses the rendering capabilities of a single NeRF trained in the entire scene, producing scene renderings of superior quality.

\section{Method}
\label{sec:method}
We first provide an overview of our DaC-NeRF to briefly discuss the workflow of our pipeline in \cref{subsec:overview}, which is followed by a detailed discussion of each component of DaC-NeRF (\cref{subsec:DaC}).
\subsection{Overview}\label{subsec:overview}
\begin{figure*}[t]
    \begin{center}
        \includegraphics[width=\linewidth]{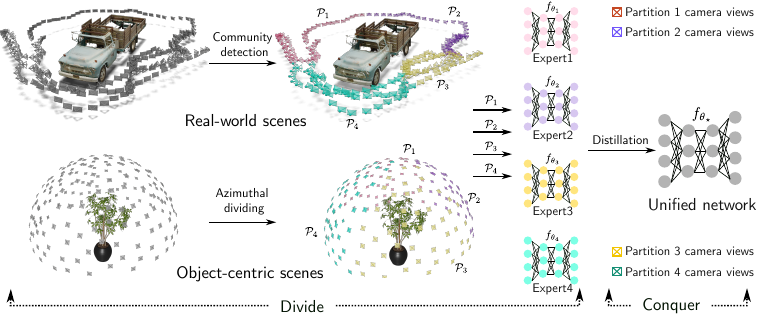}
    \end{center}
    \caption{Our Divide and Conquer (DaC) training paradigm. We illustrate our training paradigm with Truck and Ficus. In the dividing stage, we aim to partition the input views $\mathbf{P}=\{p_1,\ldots,p_N \}$ into $K$ even-sized groups $\{\mathcal{P}_1,\ldots,\mathcal{P}_K \}$. To achieve so, we propose two distinct strategies for dividing object-centric and real-world datasets based. For object-centric scenes, we divide based on the Azimuth angle of input views, while we leverage community detection in Complex Network Analysis~\cite{mohamed2019comprehensive} to divide for real-world scenes. This is followed by a teacher-student distillation to aggregate the knowledge from expert models into a unified network.
    }
\label{fig:framework}
\end{figure*}

As depicted in~\cref{fig:framework}, the workflow of our pipeline can be summarized into two stages: divide and conquer. 
In the first stage, we partition the scene into multiple subsections (refer to the color coding of~\cref{fig:framework}) such that each scene partition captures specific details of the object. 
The goal is to ensure that views within each partition share as much relevant information as possible, collectively covering all regions of the object. 
To achieve this, we design different partitioning methods for object-centric scenes and real-world 360$\degree$ scenes (see~\cref{subsubsec:divide}). 
Following the partition process, we train a specialized model (\ie, expert) on each of the partitions. This facilitates the local intricacies of the geometry to be learned by each expert. 
The subsequent distillation stage aggregates the knowledge from each expert into a single unified model; this process not only improves memory efficiency but also facilitates more efficient rendering since a single model will be employed to render new views.

\subsection{DaC-NeRF}\label{subsec:DaC}

\subsubsection{Dividing Strategy} \label{subsubsec:divide}
In this section, we introduce two different scene partition approaches for NeRF datasets. The first approach is designed for object-centric scenes, where cameras are fully described by spherical coordinates. 
The second approach is tailored for real-world 360-degree scenes, where the positions of the input cameras can be arbitrary.

\paragraph{Dividing object-centric scenes \label{para:athmu}}

We propose to divide the input camera poses $\mathbf{P} = \{p_1,\ldots,p_N\}$ into K even-sized subgroups $\{\mathcal{P}_1,\cdots,\mathcal{P}_K\}$ uniquely based on the location of their camera's centre.  
While several clustering methods based on the great-circle distance have been explored to divide the set of training views, our empirical findings indicate these methods are highly sensitive to their initial conditions, often leading to uneven clusters. In contrast, we introduce a heuristic that reliably achieves a balanced distribution of cameras across partitions, resulting in a more uniform and coherent division of the scene.

For each camera $p_i$ with associated camera centre $\mathbf{c}_i = (x_i,y_i,z_i)^T \in \mathbb{R}^3$, one can define its \textit{azimuth} coordinate given by $\phi_i=\arctan(x_i,y_i)$. The subsequent step aims at grouping the following $N$ cameras into $K$ distinct subsets $\{\mathcal{P}_1,\cdots,\mathcal{P}_K\}$ based on their \textit{azimuthal} angle. Such group allocation is given by,

\begin{equation}
    \forall \ell \in \llbracket 1, K\rrbracket, \ \  \mathcal{P}_\ell = \left\{ p_k, \frac{2\pi(\ell-1) }{K} \leq \phi_k <  \frac{2\pi\ell}{K}  \right\} .\
\end{equation}

This heuristic strategy is designed for object-centric scenes where the views' distribution is approximately uniform across the upper half-sphere of the object (refer to~\cref{subsec:implentation} for the data generation of the blender scenes).

\paragraph{Dividing real-world scenes}

To extend this method to free-form capture scenarios, it is necessary to consider extra factors related to both the camera and the scene. These include the orientation of the cameras, the geometry of the scene, and the possibility of occlusion.


It is important to emphasize that the aforementioned strategy is only effective on the fixed object-centric setup where the cameras are chosen to be uniformly distributed on a (half)-sphere. For real-world scenes, where free-form image acquisition is performed, the overlap between two camera frustums can not be described by measuring the distance between their polar angles. To extend this method to free-form capture scenarios, it is necessary to consider extra factors related to both the camera and the scene including the orientation of the cameras, the geometry of the scene, and the possibility of occlusion. \label{para:community}

To address this, we reframe this problem through the lens of community detection in Complex Network Analysis~\citep{mohamed2019comprehensive}. The goal is to group nodes in a graph so that nodes in the same group (community or cluster) are more closely connected to each other than to nodes in other groups. These algorithms often operate on graphs succinctly represented as a weighted adjacency matrix $\mathbf{A} \in \mathbb{N}^{N\times N}$ where $A_{ij}$ denotes the strength of the connection between nodes $i$ and $j$ in the graph. Note that this connection strength is application-dependent and quantifies the similarity between nodes.

In the context of scene partition, we assign a node to each view and model their connection strength using standard structure from motion (SfM) outputs. Specifically, for any pair of views $i$ and $j$, we denote their connectivity strength $A_{ij}$ as the total number of 3D points in the SfM's sparse point cloud that were triangulated from 2D feature correspondence computed on these views. We argue that this measure quantifies the similarity in visual content, field of view, and camera positioning between pairs of views. 

Once this adjacency matrix is computed, we can split the input views into $K$ subgroups based on this measure using standard graph community detection algorithms like Louvain community detection~\cite{blondel2008fast} or spectral clustering~\cite{ng2001spectral}. The former is an efficient $\left( \mathcal{O}(n\log{}n) \right)$ heuristic for optimizing graph modularity, while the latter is a clustering technique able to extract clusters with highly non-convex shapes. We encourage the reader to refer to the supplementary material for details of these algorithms.


\paragraph{Training experts}

Finally, after obtaining $K$ partitions $\mathcal{P}_1,\ldots,\mathcal{P}_K$ using either object-centric or real-world scene division strategies, we train $K$ expert NeRF models  $f_{\theta_1},\ldots, f_{\theta_K}$~\textemdash~one for each partition. This process enhances the ability of each expert to learn the local geometric nuances of their assigned scene partition more effectively.
\subsubsection{Conquering Strategy}\label{subsubsec:conquer} 
\begin{figure}[t]
    \begin{center}
        \includegraphics[width=\linewidth]{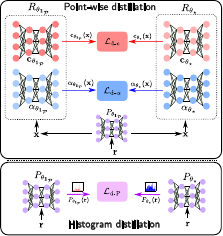}
    \end{center}
    \caption{The point-wise distillation loss introduced in our conquer strategy. At the core of our conquer strategy consists of $\mathcal{L}_{\text{d-}{\alpha}}$ to regularize $\alpha_{\theta_\star}$, $\mathcal{L}_{\text{d-}{\textbf{c}}}$ to regularize $\textbf{c}_{\theta_\star}$. Additionally, we leverage the histogram loss~\cite{barron2022mip} to guide the learning of $P_{\theta_\star}$.}
\label{fig:conquer}
\end{figure}

Maintaining $K$ expert models during the inference phase is notably inefficient. To address this, we propose merging the expertise from $f_{\theta_1},\ldots,f_{\theta_K}$ into a single integrated model represented by $f_{\theta_\star}$. This integration is achieved through a teacher-student distillation process. Without loss of generality, a NeRF model is composed of a network $P_\theta$ that samples points along a ray and a renderer network $R_\theta$ to produce the alpha compositing value $\alpha$ and color $\mathbf{c}$, which are defined by,
\begin{align}
f_{\theta_i} = \Big( P_{\theta_i} , R_{\theta_i} \Big) \\
R_\theta = (\mathbf{\alpha}_\theta,\mathbf{c}_\theta),
\end{align}
where $\alpha_\theta$ and $\mathbf{c}_\theta$ are the networks to generate alpha value and color, respectively. We introduce the indicator function to map a 3D point $\mathbf{x}$ along the ray $\mathbf{r}(t)$ into one of the expert subgroups ($f:\mathbb{R}^3 \mapsto \{1,\cdots, K\}$) by, 
\begin{equation}\label{eqn:indicator}
\mathds{1}_{\mathcal{P}_1, \cdots, \mathcal{P}_K} (\mathbf{x}) = \Big\{i \ \ | \ \ \exists t, \ \mathbf{x} \in \mathbf{r}(t) \text{ and } \mathbf{r} \in \mathcal{P}_i\Big\},
\end{equation}
where $i \in \{1,\dots, K\}$. Likewise, the same process can generalize to mapping rays into one of the expert subgroups by: $\mathds{1}_{\mathcal{P}_1,\ldots,\mathcal{P}_K}(\mathbf{r})$. For simplicity purpose, we denote $\mathds{1}_{\mathcal{P}_1, \cdots, \mathcal{P}_K}(\mathbf{x})$ or $\mathds{1}_{\mathcal{P}_1, \cdots, \mathcal{P}_K}(\mathbf{r})$ as $\mathds{1}_{\mathcal{P}}$. Having all the components defined, we can describe the distillation process hereafter.

Given a mini-batch $\mathbf{B}$ of rays $\forall \mathbf{r}\in \mathbf{B}$, $P_{\theta_{\mathds{1}_\mathcal{P}}}$ first produces a set of refined points $\mathbf{x} \in P_{\theta_{\mathds{1}_\mathcal{P}}}(\mathbf{r})$ on the rays. Subsequently, we can guide the learning of $f_{\theta_\star}$ to produce valid opacity $\alpha$ by imposing Mean-Squared Error (MSE) loss on the outputs of $\alpha_{\theta_\star}$ and $\alpha_{\theta_{\mathds{1}_\mathcal{P}}}$ (See \cref{fig:conquer}), which can be described as,
\begin{equation}
\mathcal{L}_{\text{d-}\alpha} = \sum_{\mathbf{x} \in P_{\theta_{\mathds{1}_\mathcal{P}}}(\mathbf{r})} \mathcal{L}_2 \Big(\alpha_{\theta_\star}(\mathbf{x}),\alpha_{\theta_{\mathds{1}_\mathcal{P}}}(\mathbf{x}) \Big).    \label{eqn:disalpha}
\end{equation}
In a similar fashion, the point-wise MSE loss of the color $\mathbf{c}$ can be obtained as,
\begin{equation}
\mathcal{L}_{\text{d-}\textbf{c}} = \sum_{\mathbf{x} \in P_{\theta_{\mathds{1}_\mathcal{P}}}(\mathbf{r})}  \mathcal{L}_2 \Big(\mathbf{c}_{\theta_\star}(\mathbf{x}),\mathbf{c}_{\theta_{\mathds{1}_\mathcal{P}}}(\mathbf{x}) \Big). \label{eqn:discol}
\end{equation}
Furthermore, to ensure the network $P_{\theta_\star}$ can produce valid sampled points for $R_{\theta_\star}$, we leverage the histogram loss, original proposed in~\cite{barron2022mip} to align the points generated by $P_{\theta_\star}$ and $ P_{\theta_{\mathds{1}_{\mathcal{P}}}}$. This process can be described as,
\begin{equation}
\mathcal{L}_{\text{d-P}} = \mathcal{L}_{\text{hist}}\Big( P_{\theta_\star}(\mathbf{r}), P_{\theta_{\mathds{1}_{\mathcal{P}}}}(\mathbf{r}) \Big). 
\label{eqn:disprop}
\end{equation}
Additionally, we keep the loss originally proposed in our baseline implementation and denote them as $\mathcal{L}_{\text{orig}}$ (we encourage readers to refer to the Supplementary Material for the details of $\mathcal{L}_\text{hist}$ and $\mathcal{L}_{\text{orig}}$.), such that the final loss employed in our conquer phase can be written as,
\begin{equation}
\mathcal{L}_{\text{total}} = \mathcal{L}_{\text{d-}\alpha} + \mathcal{L}_{\text{d-}\textbf{c}} + \mathcal{L}_{\text{d-P}} + \mathcal{L}_{\text{orig}}.
\label{eqn:final_loss}
\end{equation}
It is worth noting that the $\mathcal{L}_\text{orig}$ does not include the reconstruction loss between predicted colors and ground truth images since the ground truth images are not required during our conquer phase. This allows $f_{\theta_\star}$ to assimilate expertise from the experts $f_{\theta_1},\ldots,f_{\theta_K}$, within expanded space beyond the training views (interpolation between training views).
Finally, we provide the Pseudo-code of our Divide and Conquer training paradigm in~\cref{algo:divide} and \cref{algo:conquer}.

\begin{algorithm}
\caption{Divide}
\textbf{Input:} A set of posed images $\mathbf{I}=\{I_1,\ldots,I_N\}$ \\
\textbf{Output:} Expert models $f_{\theta_1},\ldots,f_{\theta_K}$
\begin{algorithmic}[1]
\If{Object-centric scenes}
    \State $\{\mathcal{P}_1,...,\mathcal{P}_K\} \gets \texttt{Azimuth\_div}(\mathbf{I})$~~~$\S1$~\cref{para:athmu}
\ElsIf{real-world scenes}
    \State $\{\mathcal{P}_1,...,\mathcal{P}_K\} \gets \texttt{Graph\_cls}(\mathbf{I})$~~~~~~~~$\S2$~\cref{para:community}
\EndIf
\For {$\mathcal{P}_i$ in $\{\mathcal{P}_1,...,\mathcal{P}_K\}$}
    \State $f_{\theta_i} \gets \texttt{Expert\_train}(\mathcal{P}_i)$ 
\EndFor
\end{algorithmic}\label{algo:divide}
\end{algorithm}

\begin{algorithm}
\caption{Conquer}
\textbf{Input:} Trained experts $f_{\theta_1},\ldots,f_{\theta_K}$ and a batch of rays $\forall \mathbf{r}\in \mathbf{B}$ \\
\textbf{Output:} Unified model $f_{\theta_\star}$

\begin{algorithmic}[1]
\State $f_{\theta_{\mathds{1}_\mathcal{P}}} \gets \texttt{Exp\_retriever}(\mathbf{r})$~~~\cref{eqn:indicator} 
\State $f_{\theta_\star} \gets \texttt{Distillation}(\mathbf{r}, f_{\theta_{\mathds{1}_{\mathcal{P}}}})$ ~~~\cref{eqn:final_loss}

\end{algorithmic}\label{algo:conquer}
\end{algorithm}












\section{Experiments and Discussion}
\label{sec:exp}
In this section, we provide implementation details of our DaC training paradigm in~\cref{subsec:implentation}. This is followed by evaluating our method on public datasets in~\cref{subsec:benchmark}. We further conduct ablation studies to justify the design choice of our proposed method in~\cref{subsec:ablation}. 

\subsection{Implementation Details}\label{subsec:implentation}
\subsubsection{Datasets} 
We train and evaluate our proposed method using NeRF synthetic datasets~\cite{mildenhall2020nerf}, including chair, ficus, hotdog, materials and ship. Hereafter, we use NeRF sysnthetic datasets and blender datasets interchangeably. 
Further, we evaluate our method on an unbounded real-world dataset, namely Tanks\&Temples~\cite{knapitsch2017tanks}, which captures 360$\degree$ of scenes, including M60, Playground, Train, and Truck.
\subsubsection*{Farthest Point Sampling} 
We have observed that insufficient quantity of training views in the original blender dataset poses a limitation in training multiple expert models, consequently failing to generalize to novel views. Moreover, the test views of the original blender scenes are distributed non-uniformly on the object, leading to a biased model in evaluation. This motivates us to opt for an alternative way to generate our training and test views for the blender scenes. 
We utilise Farthest point sampling (FPS)
for this task on the 
blender dataset, which will be discussed briefly hereafter. Given an object centred in $O \in \mathbb{R}^3$, we begin by creating a collection of candidate camera centres $\mathbf{C}=\{\Vec{x}_1,\ldots,\Vec{x}_N\}$ uniformly distributed across a hemisphere centred at the $O$ with radius $r$. Subsequently, we select a random point $\Vec{x}_j$ from $\mathbf{C}$ as the initialization. Thereafter, we compute the distances between $\Vec{x}_j$ and all other points $\Vec{x}_{i\neq j}\in\mathbf{C}$, given a predefined distance function $f$. Finally, we identify the farthest point as the next point of interest and iteratively repeat this process until $K$ points are sampled from our candidate set $\mathbf{C}$. Due to this being an object-centric dataset, all of the camera directions are pointing towards $O$, the information of the focal point alone is sufficient to discriminate between the camera's seen features. Notably, we generate $180$ training and $200$ test views respectively, across all the scenes in the blender dataset.  

\subsubsection{Hyperparameters and Loss Functions}\label{subsubsec:loss}
We follow the practice in K-planes~\cite{fridovich2023k} to train our expert models. Specifically, we train the model for $30k$ iterations with a learning rate of 0.01. Additionally, we use cosine learning rate scheduler with 512 warm-up iterations. During the conquer stage, we distill the point-wise $\alpha$ value and color $\mathbf{c}$. However, due to the two-stage ray sampling strategy (proposal-NeRF networks originally proposed in~\cite{barron2022mip}) employed in the baseline model, the distillation process becomes notably intricate. To address this challenge, we begin by acquiring the sampled point locations from the NeRF network of expert models, such that we can perform a forward pass for the NeRF network of $f_{\theta_\star}$ on these points to get the corresponding $\alpha$ and $c$. Then we impose a MSE (Mean Square Error) loss to constrain the disparities between the $\alpha$ and $\mathbf{c}$ parameters of the NeRF network of the experts and those of $f_{\theta_\star}$. Another challenge is to regularize the discrepancy of the points sampled by expert models and $f_{\theta_\star}$ since it is essentially an optimization problem on two histograms with distinct bins. To tackle this challenge, we employ the histogram loss originally imposed between the density weights of NeRF and proposal networks, proposed in~\cite{barron2022mip}. For our specific scenario, we impose the histogram loss between the $\alpha$ values produced by the NeRF network of the expert models and the proposal networks of $f_{\theta_\star}$. Note that we use the same learning rate and scheduler as training our expert models. We follow the practice in~\cite{reiser2021kilonerf} to further fine-tune the fused model $f_{\theta_\star}$ with the ground-truth images for $30k$ iterations. 
\subsection{Quantitative Results on Public Benchmarks}\label{subsec:benchmark}
We compare our proposed DaC against the standard training pipeline on the baseline model (K-Planes~\cite{fridovich2023k}) across NeRF synthetic and Tanks\&Temples datasets. We report the results on $3$ evaluation metrics, including Peak Signal to Noise Ratio (PSNR), Structural Similarity (SSIM), and Multi-Scale Structural Similarity (MS-SSIM).

\paragraph{NeRF Synthetic}

\begin{table}[tb]
    \centering
    \scalebox{0.87}{
    \begin{tabular}{l|c|c c c}
    \specialrule{.2em}{.1em}{.1em}
    
    \textbf{Scene}
    &\textbf{Model}
    &\textbf{PSNR~$\uparrow$}
    &\textbf{SSIM~$\uparrow$}
    &\textbf{MS-SSIM~$\uparrow$}
    \\

    \toprule\toprule
    \multirow{3}{*}{Chair}
    &KPlanes
    &35.05
    &0.982
    &0.996
    \\
    &KPlanes (60k)
    &35.54          
    &0.983
    &0.996
    \\
    
    &DaC-KPlanes
    &\cellcolor{red!10} 35.60          
    &\cellcolor{red!10} 0.984
    &0.996
    \\
    \hline
    \multirow{3}{*}{Ficus}
    &KPlanes
    &32.02
    &0.979
    &0.992
    \\
    &KPlanes (60k)
    &32.65        
    &0.982
    &0.993
    \\
    &DaC-KPlanes
    &\cellcolor{red!20} 32.83        
    &\cellcolor{red!10} 0.983
    &0.993 
    \\
    \hline
    \multirow{3}{*}{Materials}
    &KPlanes
    &31.95
    &0.966
    &0.986
    \\
    &KPlanes (60k)
    &32.32          
    &0.968
    &0.990
    \\
    &DaC-KPlanes
    &\cellcolor{red!10} 32.47       
    &\cellcolor{red!10} 0.969
    &\cellcolor{red!10} 0.991 
    \\
    \hline
    \multirow{3}{*}{Hotdog}
    &KPlanes
    &37.67
    &0.982
    &0.993
    \\
    &KPlanes (60k)
    &38.01          
    &0.983
    &0.994
    \\
    &DaC-KPlanes
    &\cellcolor{red!10} 38.08  
    &0.983
    &0.994
    \\
    \hline
    \multirow{3}{*}{Ship}
    &KPlanes
    &30.72
    &0.888
    &0.948
    \\
    &KPlanes (60k)
    &31.17          
    &0.894
    &0.952
    \\
    &DaC-KPlanes
    &\cellcolor{red!15} 31.29           
    &\cellcolor{red!30} 0.896
    &\cellcolor{red!30} 0.954 
    \\
    \bottomrule
    \end{tabular}
    }
    \caption{The evaluation of our DaC \vs the standard training pipeline on the blender scenes. The results demonstrate consistent improvement over the standard training pipeline across 5 different synthetic scenes (best view in color).}
    \label{tab:nerf_synthetic}
\end{table}

Based on the findings in \cref{tab:nerf_synthetic}, our DaC training paradigm demonstrates improved performance compared to the standard training pipeline across all scenes of the NeRF synthetic dataset. Notably, our DaC consistently outperforms the standard training pipeline on the baseline model, even after an additional $30k$ iterations of training. This outcome demonstrates the efficacy of our proposed DaC training paradigm for NeRF.

\paragraph{Tanks\&Temples}

\begin{table}[tb]
    \centering
    \scalebox{0.84}{
    \begin{tabular}{l|c|c c c}
    \specialrule{.2em}{.1em}{.1em}
    
    \textbf{Scene}
    &\textbf{Model}
    &\textbf{PSNR~$\uparrow$}
    &\textbf{SSIM~$\uparrow$}
    &\textbf{MS-SSIM~$\uparrow$}
    \\

    \toprule\toprule
    \multirow{3}{*}{Truck}
    &KPlanes
    &21.64
    &0.663
    &0.803
    \\
    &KPlanes (60k)
    &21.77
    &0.672
    &0.811
    \\
    &DaC-KPlanes
    &\cellcolor{red!30}22.01
    &\cellcolor{red!30} 0.702
    &\cellcolor{red!30} 0.851 
    \\
    \hline
    \multirow{3}{*}{Playground}
    &KPlanes
    &21.15
    &0.669
    &0.823
    \\
    &KPlanes (60k)
    &21.30
    &0.677
    &0.830
    \\
    
    &DaC-KPlanes
    &\cellcolor{red!10} 21.42
    &\cellcolor{red!5} 0.678
    & 0.830 
    \\
    \hline
    \multirow{3}{*}{Train}
    &KPlanes
    &17.05
    &0.574
    &0.754
    \\
    &KPlanes (60k)
    &17.23
    &0.586
    &0.766
    \\
    
    &DaC-KPlanes
    &\cellcolor{red!40} 17.66
    &\cellcolor{red!40} 0.596
    &\cellcolor{red!40} 0.773 
    \\
    \hline
    \multirow{3}{*}{M60}
    &KPlanes
    &18.43
    &0.665
    &0.741
    \\
    &KPlanes (60k)
    &18.59
    &0.872
    &0.749
    \\
    
    &DaC-KPlanes
    &\cellcolor{red!40} 19.14
    &\cellcolor{red!40} 0.686
    &\cellcolor{red!40} 0.772 
    \\
    \bottomrule
    \end{tabular}
    }
    \caption{The evaluation of our DaC \vs the standard training pipeline on the Tanks\&Temples dataset (best view in color).}
    \label{tab:tnt}
\end{table}

In this experiment, we observe a similar finding  as noted in the previous section. As ~\cref{tab:tnt} suggests, with our proposed DaC training paradigm the performance of the baseline model is improved across all the 360$\degree$ unbounded scenes, particularly on the PSNR. This consistent enhancement in rendering quality clearly shows the effectiveness of our proposed DaC.

\paragraph{Qualitative Results}
\begin{figure*}[t]
    \begin{center}
        \includegraphics[width=\linewidth]{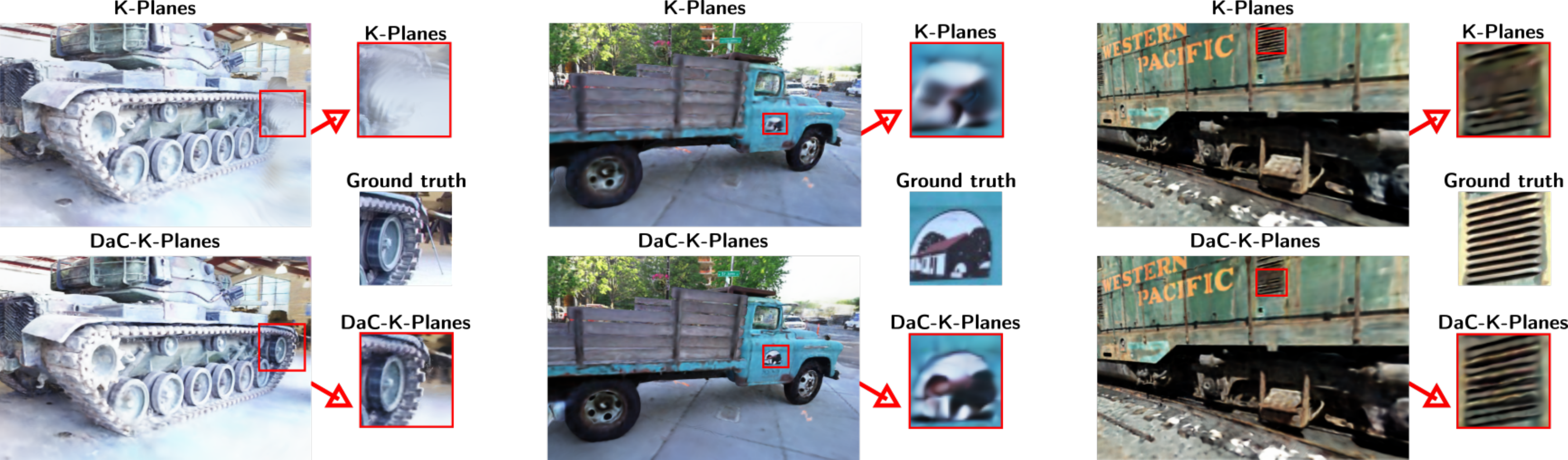}
    \end{center}
    \caption{Qualitative Results. At the $60,000$ iterations, we select the rendered test views of K-Planes, employing both the conventional training pipeline and our proposed DaC paradigm. The cropped details clearly show that our DaC training paradigm improves the rendering quality for the unbounded real-world scenes.}
\label{fig:qualitative}
\end{figure*}
For qualitative visualisation, we selected rendered test views at $60k$ iterations from the Tanks\&Temples dataset, including M60, Truck, and Train. As illustrated in~\cref{fig:qualitative}, the conventional NeRF training pipeline exhibits limitations in the nuances of the captured details within a fixed training budget. In contrast, our proposed DaC clearly improves rendering quality, a distinction particularly evident in the M60 and Truck views, which clearly shows the superiority of our proposed DaC training paradigm.

\subsection{Ablation Study}\label{subsec:ablation}
In this subsection, we conduct various ablation studies to verify our design choices, including investigating the number of partitions, assessing performance with or without overlap between partitions, assessing different partitioning methods for Tanks\&Temples, and finally assessing the number of distillation and fine-tuning steps.  

\subsubsection{The Number of Partitions}\label{subsubsec:partitions}
\begin{table}[tb]
\centering
\scalebox{0.92}{
\begin{tabular}{c|ccccc|c}
\specialrule{.2em}{.1em}{.1em}
 & \textbf{Chair} & \textbf{Ficus} & \textbf{Hotdog} & \textbf{Materials} & \textbf{Ship} &\textbf{Avg} \\
\toprule\toprule
2 & 35.52 & 32.80 & 38.01 & 32.43 & 31.21 & 33.39\\
3 & \bf{35.61} & 32.83 & 38 & 32.51 & 31.22 &34.03\\
\cellcolor{red!10}4 & \cellcolor{red!10}35.60 & \cellcolor{red!10}\bf{32.83} & \cellcolor{red!10}\bf{38.08} & \cellcolor{red!10}32.47 & \cellcolor{red!10}\bf{31.29} & \cellcolor{red!10} \bf{34.05}\\
5 & \bf{35.61} & 32.82 & 38.04 & \bf{32.52} & 31.24 &34.05\\
\bottomrule
\end{tabular}
}
\caption{The analysis of the number of partitions. We conduct the experiments on the NeRF synthetic scenes. Note that the evaluation is reported in PSNR.}
\label{tab:num_partition}
\end{table}

In this ablation, we investigate how the number of partitions effect the final result.
We conduct experiments on NeRF synthetic scenes with the Peak Signal to Noise Ratio (PSNR) evaluation metric. We compare the test performance across 4 partitioning setups (\ie, 2-5). The results in~\cref{tab:num_partition} suggest that dividing the views into 4 partitions yield the best trade-off between performance and efficiency as increasing the number of partitions over $4$ yields no further significant performance gain.

\subsubsection{Overlap Between Partitions}
\begin{figure}[t]
    \begin{center}
        \includegraphics[width=\linewidth]{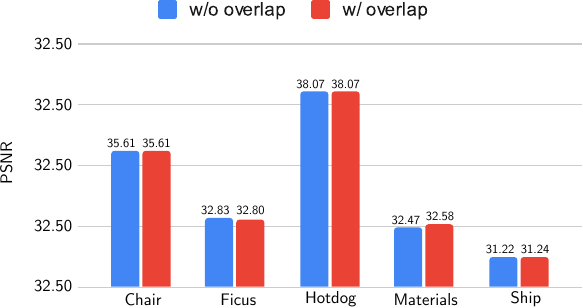}
    \end{center}
    \caption{
    The ablation study on overlapping partition views. In this experiment, we train 6 NeRF models with 60$\degree$ overlapping azimuth angle. The results demonstrate that training overlapping NeRF models with our DaC yield no significant performance gain.}
\label{fig:overlap}
\end{figure}

We investigate the efficacy of training a dedicated NeRF model on the boundary region between two partitions. This experiment is conducted using the NeRF synthetic scenes, employing the Peak Signal-to-Noise Ratio(PSNR) as the evaluation metric. Specifically, we train a total of six expert models on the boundaries, each covering 120$\degree$ azimuthal angle with a 60$\degree$ overlap. The results in~\cref{fig:overlap} indicate that such an approach does not yield obvious performance improvements, which is consistent across all scenes.


\subsubsection{Distillation and Fine-Tuning Steps}
\begin{table}[tb]
\centering
\scalebox{0.8}{
\begin{tabular}{cc|ccccc|c}
\specialrule{.2em}{.1em}{.1em}
\textbf{DST} &\textbf{FT} &\textbf{Chair} &\textbf{Ficus} &\textbf{Hotdog} &\textbf{Materials} &\textbf{Ship} &\textbf{Avg}\\
\toprule\toprule
10k & 50k & 35.63 & 32.76 & 36.10 & 32.52 & 30.44 & 33.49\\
20k & 40k &\bf{35.63} &32.80 &38.00 &\bf{32.56} &31.14 & 34.03\\
\cellcolor{red!10}30k & \cellcolor{red!10}30k &\cellcolor{red!10}35.60 &\cellcolor{red!10}\bf{32.83} &\cellcolor{red!10}\bf{38.08} &\cellcolor{red!10}32.47 &\cellcolor{red!10}\bf{31.29} &\cellcolor{red!10} \bf{34.05}\\
40k & 20k & 35.43 & 32.60 & 37.87 & 32.18 & 31.10 & 33.83\\
50k & 10k & 35.01 & 31.70 & 37.16 & 31.42 & 30.52 & 33.16\\
\bottomrule
\end{tabular}
}
\caption{The ablation study of the effect of distillation \vs fine-tuning. Note that the evaluation is reported in PSNR. The results clearly suggest that a balanced distillation (DST) and fine-tuning (FT) iterations (30k for each) yields the best result.}
\label{tab:ablation_distil_fine_tune}
\end{table}

Per the findings outlined in~\cref{subsubsec:partitions}, it is evident that dividing the input views into four groups showcases the best trade-off between local specialization and computational budget. Nevertheless, there remains a necessity to investigate the capacity of how such a training framework could be further enhanced by varying the distillation and fine-tuning with a fixed $60k$ iterations of training budget. We conduct this experiment on NeRF synthetic scenes using PSNR as the evaluation metric. In~\cref{tab:ablation_distil_fine_tune}, we observe that a balanced iterations between distillation and fine-tuning (denoted as \textbf{DST} and \textbf{FT} in~\cref{tab:ablation_distil_fine_tune}) yield the optimal performance.

\subsubsection{Discussion}
Our primary interest focus is to mitigate the limitations inherent in the conventional NeRF training pipeline, specifically in the context of synthetic and unbounded $360\degree$ scenes. We demonstrate improved consistent performance over our baseline model for novel view  synthesis. The ablation study verifies our design choice.  
\section{Conclusion and Future Work}
\label{sec:conclusion}
In this research, we introduce a pioneering training paradigm for NeRFs, inspired by ensemble based approaches for improving novel view synthesis. Our approach provides insights on how input views of a real-world scene can be effectively partitioned into multiple subgroups, to facilitate the learning of spatial expert models. Our findings demonstrate that these spatial experts achieve superior rendering quality on their local regions. We further showcase the potential of knowledge distillation to aggregate the expertise into a unified model, reducing memory complexity dramatically during the inference stage. Our proposed DaC exhibits frame-work characteristics, making it adaptation-friendly to scene representation methods that may surface in the future.  

In future work, we wish to extend such an idea to dynamic scenes. In a similar fashion, one can decompose the time axis into multiple segments and models can learn to specialize on different time partitions such that the final learned model would exhibit better spatial-temporal consistency. Furthermore, another potential extension could be applying our DaC approach into a continual learning setup where the training data is presented in a streaming manner.

{
    \small
    \bibliographystyle{ieeenat_fullname}
    \bibliography{main}
}


\end{document}